# A Pulse-and-Glide-driven Adaptive Cruise Control System for Electric Vehicle

Zhaofeng Tian[1], Liangkai Liu[2], and Weisong Shi[3]

*Abstract*— As the adaptive cruise control system (ACCS) on vehicles is well-developed today, vehicle manufacturers have increasingly employed this technology in new-generation intelligent vehicles. Pulse-and-glide (PnG) strategy is an efficacious driving strategy to diminish fuel consumption in traditional oil-fueled vehicles. However, current studies rarely focus on the verification of the energy-saving effect of PnG on an electric vehicle (EV) and embedding PnG in ACCS. This paper proposes a pulse-and-glide-driven adaptive cruise control system (PGACCS) model which leverages PnG strategy as a parallel function with cruise control (CC) and verifies that PnG is an efficacious energy-saving strategy on EV by optimizing the energy cost of the PnG operation using Intelligent Genetic Algorithm and Particle Swarm Optimization (IGPSO). This paper builds up a simulation model of an EV with regenerative braking and ACCS based on which the performance of PGACCS and regenerative braking is evaluated; the PnG energy performance is optimized and the effect of regenerative braking on PnG energy performance is evaluated. As a result of PnG optimization, the PnG operation in the PGACCS could cut down 28.3% energy cost of the EV compared to the CC operation in the traditional ACCS which verifies that PnG is an effective energy-saving strategy for EV and PGACCS is a promising option for EV.

## I. INTRODUCTION

THE conception of adaptive cruise control (ACC) could be traced to 1960s [1], which immediately gained much attention from America transportation departments. Continuously developing in past several decades, the ACC now could meet the needs of car-following function both in highway and urban operating conditions [2, 3] and considerably reduce the driving burden for drivers [4]. Following the ACC function becoming fairly mature, the speed control strategy and space control strategy for adaptive cruise control system (ACCS) are branching to several approaches like PID algorithm [5, 6], fuzzy algorithm [7, 8] and model predictive control (MPC) [2, 9] to control the vehicle speed; There are also constant time headway [10] strategy and variable time headway [11] strategy as space control strategies to keep appropriate space from the leading vehicle. The widely accept ACCS model consists of two subsystems, one is ACC subsystem to implement ACC function and the other one is cruise control (CC) system to implement CC function [2, 13]. This study selects this model that uses a combination of ACC and CC subsystems to form the ACCS.

In recent years, the researches on ACCS of EV have been reported. [12, 13, 14] used MPC algorithms to meet the requirements of ACCS on electric vehicle (EV). Although some researches mentioned the advantages of EV on energy consumption, all of them naturally focus on the implementation of control strategies but rarely the regenerative braking and energy-saving strategies in lack of the modelling of regenerative braking and exploiting the energy problems.

Pules-and-glide (PnG) is an energy-saving strategy which has been verified a more efficient approach to save fuel based on gasoline fueled vehicle than conventional cruise control cruising at a constant speed [15]. To improve the comfort of passengers, [16] proposed a smoother PnG approach based on a hybrid vehicle model constrained with a certain gear ratio. But no paper discusses the PnG strategy and related problems on an EV based on a pure EV model.

According to above-mentioned studies, PnG strategy is rarely discussed on electric vehicles researching that does PnG an effective energy-saving approach to apply on EV. Rarely also the PnG is considered as an embedded function in ACCS. The PnG function in an ACCS, however, could be potentially an alternative mode parallel to CC mode that could substitute CC mode to curtail the energy consumption when EV battery state-of-charge (SOC) stays at a low level. Besides the above two main research gaps, as an important EV function and an energy-saving method, the regenerative braking is scarcely considered and discussed in ACCS or PnG studies. The regenerative braking plays a significant role in EV because it could recover a part of energy loss of the braking operation, transforming partial kinetic energy to electrical energy to charge the battery pack of EV. In this case its effect on EV and PnG operation is necessary to be well

Z. Tian, L. Liu, W. Shi are with the College of Engineering, Wayne State University, Detroit, MI 48202, USA (e-mail: gy8217@wayne.edu)

discussed. This paper develops a comprehensive consideration of these problems to bridge the research gap.

To obtain the best energy-consuming performance of PnG strategy, it is necessary to optimize the parameters of PnG strategy. Using an optimization method to solve PnG issues are one of the originalities in this study. In the research domain of optimization algorithms, some heuristic algorithms were proved of high efficiency in engineering applications. Two of them are the genetic algorithm (GA) and the particle swarm optimization (PSO), GA and PSO are both popular search algorithms seen used in wide-scale researches while GA places attention on the natural search of optimization [34] and PSO emphasis on the comparative process search [35]. After the two successful algorithms, a mixed algorithm is getting attention, the intelligent genetic algorithm and particle swarm optimization (IGPSO), which is the one this study uses to optimize the PnG parameters. IGPSO is proved capable of taking advantage of the merits from GA and PSO by simultaneously taking advantage of powerful global search of GA and fast increasing optimized swarms in PSO that increases the computation efficiency [18]. Besides these classical optimization algorithms, other interesting algorithms are proposed recently. Paper [36] designed a robust PID controller based on the whale optimization algorithm to stabilize the wind energy system against wind speed fluctuations. [37] proposes a new variable structure gain scheduling (VSGS) for frequency regulation. The VSGS can leverage the merit of both the BFOA-based PI controller and GA-based PI controller to minimize both the settling time and the overshoot. [38] proposes an optimal design for the nonlinear model predictive control (NLMPC) based on modified multitracker optimization algorithm (MMTOA) to track different linear and nonlinear trajectories on robotic manipulator. [39] uses crow search algorithm to optimize the parameters of the adaptive model predictive control for blade pitch control in wind energy system.

This paper is devoted to evaluating the effect of PnG's energy-saving capability on an EV by the modelling of ACCS on EV with regenerative braking based on which energy problems are formulated. This paper also considers PnG an embedded mode or function parallel to cruise control (CC) function in an ACCS based on which this study proposes the pulse-and-glide-driven adaptive cruise control system (PGACCS). As far as specific processes are concerned, the first problem of this study is to:
- Builds up the EV and PGACCS model and validate ACC function in the PGACCS.

Besides the problem of the implementation of ACC function in the PGACCS, this paper considers formulates three related energy problems on the EV and PGACCS model:

- The effect of regenerative braking on SOC recovery in a certain driving cycle;
- The optimization of the PnG strategy employed by EV with a certain optimization algorithm.
- Evaluating the effect of regenerative braking on the PnG operation.

To solve the four proposed problems, this study designs various simulation scenarios or operating conditions corresponding to each problem, the details of which could be described as follows.
- In the first problem, the implementation of ACC function in the PGACCS, two scenarios including the leading vehicle accelerating and the other vehicle cutting in line are designed to test the vehicle-following capability;
- The second problem is about energy problem on regenerative braking embedded in the EV model, in which the energy consumption of this function is tested in an New European Driving Cycle (NEDC) that consists of urban and highway driving conditions and is widely accepted by manufactures to evaluate energy consumption performance of a vehicle;
- The third problem is an energy problem too, it is committed to exploit the lowest cost in the PnG operation within the constrain of speed fluctuation and demonstrate the energy-saving advantages compared to cruise control (CC). To minimize the energy cost in the PnG operation by finding the optimal value of acceleration of the pulse phase and deceleration of the glide phase and then this study compares the optimized PnG energy cost with the CC operation demonstrating the energy-saving advantages of the PnG operation.
- The fourth problem is evaluating the energy effect of regenerative braking on the PnG operation. In this problem, this study imposes regenerative braking in the PnG operation by assigning a value to the deceleration of the glide phase that triggers regenerative braking according to the EV dynamic model. Then this study compares the energy cost with the optimized PnG operation to find the energy-consumption results of imposing generative braking in the PnG operation.

In summary, this paper makes the following key contributions from above-mentioned problems shown as follows:
- This study build a model for the adaptive cruise control system on an EV with regenerative braking, which is rarely touched on in other studies.
- This study discusses the energy problem brought up by regenerative braking function embedded in the EV model both in NEDC driving and pulse-and-glide driving.
- This study proposes an optimization method based on IGPSO for the optimal PnG strategy that demonstrates the energy-saving effect of

PnG operation on EV. This key contribution bridges the research gap in the EV and ACCS field.

The organization of this paper can be described as follows. In Sec. II, the background of relative researches and the motivation of this paper are introduced. The electric vehicle model and PGACCS model are respectively present in Sec. III and Sec. IV. The problems including one ACC problem and three energy problems are presented in Sec. V With their solutions. The simulation study and the evaluation of its results are shown in Sec. VI. The discussion about the study and PnG is in Sec. VII. The conclusion based on simulation results and discussion is drawn in Sec. VIII.

## II. BACKGROUND AND MOTIVATION

### A. ACCS and EV

Currently the intelligent vehicles are paid increasingly attention to deal with the traffic accidents. Based on the improvement of sensing technology and the modern automotive industry, the market developed consuming interests in diver assistance functions on vehicles thus advanced driver assistance systems (ADAS) are developed and employed to produce the more advanced intelligent vehicles [18]. With the increasing ownership of fossil oil fueled vehicles, the problems of environmental pollution and oil shortage are getting sharper [19]. To solve the consequent problems brought up by traditional oil fueled vehicles, electric vehicle (EV) is a good solution to decline the air pollution [14, 20]. When doingv the research on EV, one thing could not be overlooked is that the configuration and characteristics of EV have discrepancy with traditional vehicles [13, 21, 22, 23]. For example, the energy density of EV is lower than oil fueled vehicles due to the energy density restrains on battery packs that make EV require higher energy performance. In this case, the energy problems on EV should be paid enough attention to. When designing the ADAS function on an EV, the energy cost should be evaluated as a significant course.

The adaptive cruise control (ACC) is an important function among various ADAS technologies and is an extension of traditional cruise control (CC), which is used to improve the driving comfort and relieve the driving burden for drivers [24]. When employing adaptive cruise control on a vehicle, the vehicle could receive information of distance and relative velocity to the leading vehicle from sensors like Laser and Radar, then it could adjust the speed and space to the leading vehicle based on the embedded spacing strategy to keep a safe distance as well as maintain the following capacity [25]. In some researches, adaptive cruise control and cruise control are both regarded as functions under an adaptive cruise control system (ACCS) [2, 13]. That is because when there is no leading vehicle, adaptive cruise control has no object to follow, thereby the ego vehicle needs to switch to the cruise control mode. As a result, adaptive cruise control system should combine ACC and CC together to deal with complex road conditions. This study also selects this approach to develop the vehicle model regarding ACC and CC both functions under the adaptive cruise control system (ACCS).

As an important component of ACCS, controller plays a significant role in making system accurate enough in keeping appropriate distance to the leading vehicle. In current researches, controller designing has drawn much attention that there are various control algorithms are employed in the designing process. For instance, there are sliding model control [26], reinforcement learning [27], fuzzy control [7, 8], PID control [5, 6] and model predictive control [2, 9]. Model predictive control is widely used in industries and has advantages of prediction for a real-time system which gives itself a position on automotive industry. PID controller is also a good selection for ACCS that benefits from its simple design process and sufficient accuracy and short computation time.

Besides above-mentioned researches on the ACCS and EV. Recently there are also some researches study on the ACCS of EV. Literature [12, 13, 14] employs model predictive control algorithm to implement the ACC function. But these studies naturally focus on control algorithms, which means the bodies of these studies are implementation of ACC function without presenting the EV characteristics like EV modelling especially the regenerative braking system and battery SOC model.

### B. Regenerative Braking System

As for regenerative braking system on EV, there are some researches aiming to enhance its performance. Literature [28] proposed a braking force distribution based on the ideal distribution curve. [29] proposed the genetic algorithm to increase the braking energy recovery and maintain vehicle stability. [30] designed a sliding mode controller to coordinate the antilock brake force and the regenerative brake force. All these researches focus on improving the performance of regenerative system itself, although some may discuss energy problems, still in lack of comprehensive discussions that not only consider the regenerative braking but also the EV.

### C. Researches of Energy-saving Approaches on EV

Currently the researches of energy-saving approaches are mainly focusing on the hardware-level components. Follows are some researches about energy-saving approaches on EV. Literature [31] describes the application of the electric double layer capacitor (EDLC) blocks with the change-over between the series and the parallel modes, to the traction inverter circuit for pure electric vehicles that could improve energy density on EV. Literature [32] proposes a novel load management solution for coordinating the charging of multiple plug-in electric vehicles (PEVs) in a smart grid system to diminish power loss in charging. In literature

[33] a fuzzy based safe power management system for EV is proposed to increase the effectiveness of energy-saving in terms of speed and acceleration. These researches evaluate the approaches on EV's energy-saving from the consideration of energy reservation unit, charging grid and energy management. These researches focused on improving EV's energy performance with the certain hardware technologies.

*D. PnG strategy*

A strategy for energy-saving through driving maneuver is called pulse-and-glide (PnG) strategy, which consists of two phases, the pulse phase and the glide phase. In the pulse phase, the vehicle accelerates through pressing the accelerator pedal and in the glide phase, the vehicle decelerates through sliding or braking. This strategy is an efficacious energy-saving strategy that has been proved a more efficient approach based on gasoline fueled vehicle to save fuel than conventional cruise at a constant speed or cruise control (CC) under ACCS [15]. To improve the comfort of passengers, literature [16] proposed a smoother PnG approach based on a hybrid vehicle model constrained with a certain gear ratio and a special road condition. However, there is no research talking about the PnG operation on EV and exploiting its energy performance on EV where the EV characteristics like regenerative braking would affect its performance and make the situation different from the traditional or hybrid vehicles.

*E. Motivation*

According to above-mentioned researches, the principal research gaps can be summarized as follows:
- No study considers PnG a potential function under ACCS and build a corresponding model based on it.
- No study employs PnG strategy on EV as a method to cutting down energy consumption and exploit the optimization of PnG strategy on EV as well as the effect of EV's features on the PnG operation.

To bridge these research gaps, associating PnG with ACCS and exploiting energy improvements on it as well as the energy effect of EV characteristics on the PnG operation, this paper proposes an PGACCS model of EV that contains PnG strategy as a parallel function of cruise control in the ACCS. Moreover, this study exploits the energy problems like the energy-saving effect of regenerative braking of EV in a driving cycle and the optimization of PnG strategy to achieve lower energy cost and evaluates the energy effect of regenerative braking on PnG. The topics of related works and this paper as well as the relationships among them are shown as Fig.1.

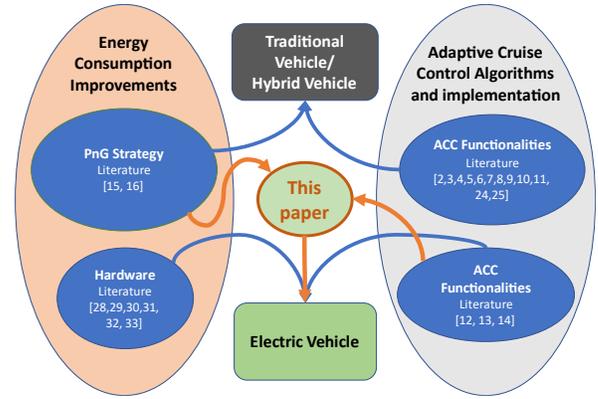

Fig. 1. Topics of related works and this paper

III. ELECTRIC VEHICLE MODELING

The vehicle model is selected as D-class sedan, selected in Carsim. Due to all vehicle models in the software are fuel vehicle, this study disconnects the fuel powertrain and replace it with an electric powertrain including a motor model and a battery model in the Simulink. Also, the PGACCS is built in Simulink. To simplify the modelling of this study, the main structure of the EV with PGACCS is presented in Fig. 2, where the PnG mode is paralleled to the cruise control mode (CC mode) in order to evaluate the effect of PnG strategy. The work of this section is modelling the motor and battery and providing the determined parameters for the EV.

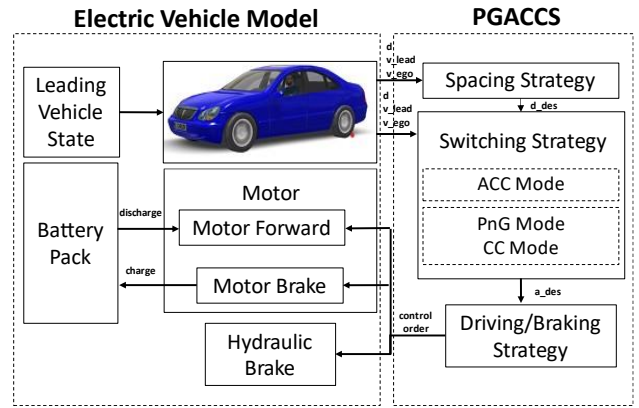

Fig. 2. Electric vehicle model with ACC system

*A. Vehicle Model*

This vehicle is a D-class sedan with electric powertrain. The key parameters of the vehicle invoked by simulation are presented in Table I.

TABLE I. VEHICLE PARAMETERS

| Parameter | Description | Value | Unit |
|---|---|---|---|
| M | vehicle mass | 1458 | kg |
| $C_d$ | air drag coefficient | 0.33 | – |
| $C_r$ | rolling resistance coefficient | 0.012 | – |
| $A_f$ | vehicle frontal area | 2.3 | $m^2$ |
| $\rho_a$ | air density | 1.206 | Kg/m$^3$ |
| g | acceleration of gravity | 9.81 | m/s$^2$ |
| ig | gear ratio (1-5) | [ 3.62, 1.93, 1.29, 0.93, 0.69] | – |
| $\eta_r$ | gear efficiency (1-5) | [0.92, 0.92, 0.95, 0.95, 0.98] | – |
| $i_f$ | final drive ratio | 3.87 | – |
| $r_w$ | effective tire radius | 0.33 | m |
| $C_b$ | battery capacity | 25 | A.h |

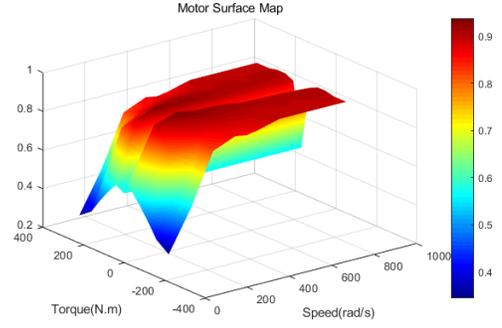

(a) Motor surface map

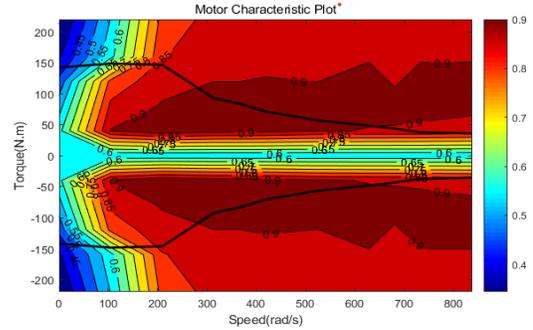

(b) Motor surface map

Fig. 3. Motor characteristics

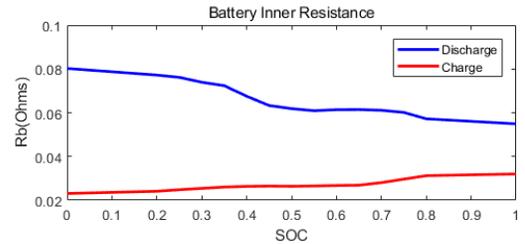

(a) Battery inner resistance

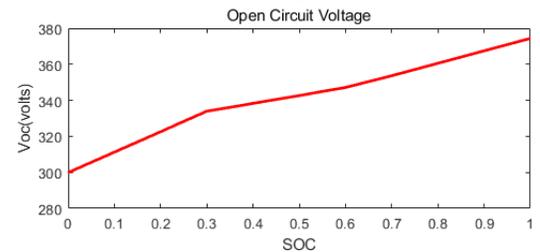

(b) Open circuit voltage

Fig. 4. Battery characteristics

## B. Electrical Powertrain Model

The motor characteristic diagram and surface map are shown as Fig. 3, in which the max torque is 150 N.m as two black lines indicate in the characteristic diagram. The inner resistance and open circuit voltage of the battery pack on this EV are presented by Fig. 4.

As for SOC calculation, this study selects commonly used open-circuit-voltage-resistance (OCV-R) model, and uses current-integration method to calculate the SOC:

$$SOC = \frac{C_{left} - \int_0^t i_b(t)dt}{C_b} \quad (1)$$

$$I_b = \frac{V_{oc} - V_{oc}\sqrt{1 - 4P_b R_b / V_{oc}^2}}{2R_b} \quad (2)$$

Where $I_b$ is the battery current, $C_{left}$ is the battery capacity left and $C_b$ is the nominal battery capacity. $P_b$ is the battery power and could be described as below:

when battery is charging,

$$P_b = \eta_m P_m \quad (3)$$

when battery is discharging,

$$P_b = \eta_m^{-1} P_m \quad (4)$$

where $\eta_m$ is the efficiency of the motor and Pm is the power of the motor which can be obtained from Fig. 3.

## C. Dynamic Model

To simplify the simulation reasonably, this study only considers rolling resistance force and longitudinal aerodynamic force on the flat ground with no slope. The dynamic model can be written as:

$$F_{dri} = C_r M g + C_d \rho_a A_f V_{rel}^2 / 2 + M_e a_{des} \quad (5)$$

$$T_m = \frac{F_{dri} r_w}{i_g i_f} \quad (6)$$

Where $F_{dri}$ is the total driving force of vehicle; $V_{real}$ is the vehicle speed relative to wind speed. Assuming the wind speed is 0 that $V_{real}$ is the real vehicle speed; $M_e$ is the equivalent mass of vehicle, which in this study is selected as $M_e=1.1M$. In the ACC system, the order of acceleration is generated by speed controller and finally executed by the motor, the relation between desired acceleration $a_{des}$ and desired motor torque $T_m$ can be illustrated by equation 5 and 6. Then the desired motor power could be obtained from equations below:

$$P_m = T_m \omega_m \quad (7)$$

$$\omega_m = V_{real} r_w i_g i_f \quad (8)$$

where $T_m$ is the torque of the motor and $\omega_m$ is the rotation speed of the motor. and then $P_m$ can be used to calculate the SOC.

## D. Brake system

The brake system of this EV consists of motor brake (regenerative braking) and hydraulic brake as a simplified brake system which do not consider ABS or slip. When desired deceleration is generated in the speed controller, the brake torque on the wheels will be distributed to the motor and hydraulic brake system which could be described as

$$M_e a_{dd} = C_r M g + C_d \rho_a A_f V_{real}^2 / 2 + F_{brk} \quad (9)$$

Where $a_{dd} > 0$ is desired deceleration, $V_{real}$ is vehicle speed and $F_{brk}$ is total brake force. In order to gain more SOC recovery, this study intends to employ regenerative brake as more as possible so that only when the max motor brake torque is below the desired brake torque, the hydraulic brake will be engaged. When only motor brake employed, there is:

$$T_{mb} = \frac{F_{brk} r_w}{i_g i_f} \quad (10)$$

where $T_{mb}$ is the motor brake torque whose curve could be obtained from Fig.3 where the motor speed is negative.

As for hydraulic brake system. The relation between hydraulic brake torque and pressure on four wheels could be written as

$$T_{hb} = C_{brk} P_b \quad (11)$$

where $T_{hb}$ is the brake torque of hydraulic brake system and $P_b$ is the total brake pressure on four wheels while the $C_{brk}$ is torque/pressure coefficient selected as 1500 on this EV. The braking strategy is described in Sec. III.

## IV. PGACCS MODELING

This section presents the modelling of PGACCS that contains its basic components like spacing strategy, speed controller, mode switch strategy, driving/braking strategy and PnG strategy.

### A. Spacing Strategy

The spacing strategies based on time headway could be generally divided into variable-time-headway (VTH) strategy and constant-time-headway (CTH) strategy. VTH strategy introduces differentiation element into spacing control to make the system response faster to the distance error, while also amplify the system error and measurement noise [11]. For a robustness reason, this study adopts CTH strategy and it can be presented as

$$d_{des} = d_0 + t_h v_{lead} \quad (12)$$

$d_{des}$ is the desired distance to the leading vehicle, $t_h$ is the time headway and $v_{lead}$ is the speed of the leading vehicle. In this study, $t_h$ is set to 1.5 seconds and $d_0$ is set to 2 meters.

### B. Speed Controller

This study does not focus on the control algorithms thus it selects PI controller for vehicle speed control, which has intelligible mechanism and strong robustness. The input of PI controller is the error between the preset ego vehicle speed $V_{des}$ and real ego vehicle speed $V_{real}$ and the output is the desired acceleration. And the PI controller model could be described as following equations

$$e(t) = V_{des}(t) - V_{real}(t) \quad (13)$$

$$u(t) = K_p e(t) + \frac{1}{T_i} \int_0^\tau e(t) dt \quad (14)$$

$$G(s) = \frac{U(s)}{E(s)} = K_p \left(1 + \frac{1}{T_i} s\right) \quad (15)$$

where $K_p$ is the proportionality coefficient; $T_i$ is the integral time constant.

### C. Mode Switching Strategy

The switching strategy of adaptive cruise control mode and cruise control mode or PnG mode is defined in Table II and equations (16) and (17)

TABLE II. SWITCHING STRATEGY

| Distance | Vp>Vset | Vp≤Vset |
|---|---|---|
| d<$d_{des}$ | ACC | ACC |
| $d_{des}$≤d<$d_{logic1}$ | CC or PnG | ACC |
| $d_{logic1}$≤d<$d_{logic2}$ | CC or PnG | Buffer |
| d≥$d_{max}$ or d≥$d_{logic2}$ | CC or PnG | CC or PnG |

$$d_{logic1} = d_{des} + k_1(V_{des} - V_{lead}) + k_2(V_{real} - V_{des}) + d_1 \quad (16)$$

$$d_{logic2} = d_{des} + k_3(V_{des} - V_{lead}) + k_4(V_{real} - V_{lead}) + d_2 \quad (17)$$

where $d_{logic1}$ is the distance between the ego vehicle and the leading vehicle when the leading vehicle speed $V_{lead} \leq V_{des}$ and the system switches from CC/PnG to ACC. $d_{logic2}$ is the distance between the ego vehicle and the leading vehicle when $V_p \leq V_{set}$ and ACC system switches from ACC to CC or PnG. $d_{max}$ is the largest range of sensors, this study selects $d_{max}$=200m. $k_1$ is relative to $V_{des}$-$V_{lead}$, which after fitting, could be described as:

$$k_1 = 1.999 - 1.196e^{-0.1299(V_{des} - V_{lead})} \quad (18)$$

This study selects $k_2 = 1.2$, $k_3$=2.9, $k_4$=1.25, $d_1$= 2m, $d_2$= 2m based on the experience of experiments.

### D. Driving/Braking Strategy

Assuming the resistance force consists of just rolling resistance force and aerodynamic force. So that the sliding acceleration could be written as

$$a_s = -\frac{1}{M}\left(C_r M g + C_d \rho_a A_f V_{real}^2 / 2\right) \quad (19)$$

And the driving/braking strategy could be described as

$$\begin{cases} a_{des} > a_s + a_0 & Drive \\ a_{des} < a_s - a_0 & Brake \\ else & Slide \end{cases} \quad (20)$$

where $a_0$ is a buffer set to avoid frequent switching between driving mode and braking mode. This study selects $a_0 = 0.05$. To the safety and comfort reasons, this study set a limitation of maximum acceleration at 2 m/s² and an limitation of maximum deceleration at -2 m/s².

### E. Pulse-and-Glide Strategy

PnG strategy is an efficient approach to cut down the energy consumption, however, its operating condition is constrained by safety and comfort requirements based on practical experience. In the car-following scenario where adaptive cruise control works, the safety factor should be paid most attention on. Although theoretically the ego car is able to execute PnG strategy while not violate the minimum safety space to the leading vehicle preset in the spacing strategy of ACC system in a car-following scenario, the constraints of the operating condition is strict that the leading vehicle should maintain a constant speed [16]. Thus, executing PnG operation in an ACC process could not match the common concern of ACC because the ability of tracking following the leading car closely is impacted. What is more, the safety cost should not be overlooked that the vehicle following the ego vehicle would accelerate and brake more frequently or increase the front space to the ego car, which would do harm to transportation efficiency and road safety.

Due to the above-mentioned reasons, this study just settles the PnG strategy in a cruise scenario where the CC mode works in a traditional ACC system. In PGACCS, the PnG performs as mode parallel to CC mode which means when the road condition allow vehicle to cruise (no leading car or leading car far away), driver could select PnG or CC mode depends on their preference and SOC because PnG could be an alternative option to cut down SOC consumption saving more energy than CC ( demonstrated in Sec. V).

## V. PROBLEM FORMULATION AND SOLUTIONS

In this section, four problems are formulated, they are ACC function implementation, regenerative braking performance, PnG energy consumption optimization and the effect of regenerative braking on PnG. Their solutions are also respectively proposed in this section.

### A. ACC Function implementation

The first objection of this section is to implement ACC function of PGACCS on EV and evaluate the performance of ACC function built in the Sec. III. In this case, two specified car-following scenarios are designed to assess the car-following capabilities of the ACC function.

The first scenario could be described as follows. When t < 30s, the leading car cruise at 60 km/h and then accelerate to 90km/h with an acceleration of 1.6 m/s². The ego car follows the leading car with an initial position of $d_{des}$ behind the leading car and an initial speed of 30 km/h.

The second scenario could be described as follows: when t < 15s, the leading car cruise at 80 km/h and the ego car follows the leading car with the inter-car distance of $d_{des}$. Then when t = 15s, another car cut in from the other lane 20m ahead of the ego car.

Three parameters are selected to evaluate the performance of the following capabilities, they are the displacement of the leading vehicle and the ego vehicle, the speed of the two vehicles and the inter-car distance. If the ACC function works well, the difference of the displacement of the two vehicles should be fairly stable, the speed of two vehicles should finally be same when no perturbation imposed and the inter-car distance should be at preset inter-car distance according to the spacing strategy.

### B. Regenerative Braking Performance

In this part, the objective is to figure out how does regenerative braking affect energy consumption in a certain driving cycle. Which means how much energy can be saved by regenerative braking in the driving cycle and how much is the effect of it compared with the situation where the regenerative braking is not used and the braking commands are solely executed by hydraulic braking.

To answer the question above, this study employs the New European Driving Cycle (NEDC) to carry out the simulation study for testing the energy consumption between employing regenerative braking and solely employing hydraulic braking and presents their energy performance in a contrastive way.

To evaluate the energy performance of regenerative braking, the SOC cost of battery is regards as a parameter. If the performance is satisfactory, the SOC cost of employing regenerative braking should be lower than employing hydraulic braking only.

## C. PnG Energy Consumption Optimization

PnG is verified an effective approach to cut down the fuel consumption on gasoline vehicles and hybrid vehicles. The mechanism of saving energy in a gasoline fueled vehicle is basically constraining the engine to perform on the high-efficiency zone [15][16]. In a hybrid vehicle, the motor could engage when engine is not efficient enough and lower the fuel consumption of engine by smoothing the curve of engine power output vs fuel rate [16]. Although the fuel consumption is lowered, the total consumption including fuel consumption and electricity consumption are not mentioned which should be concerned when energy problem is the interest of the research.

Based on the results of former PnG researches on gasoline or hybrid vehicles, this study will research on the energy consumption problem of the PnG operation on a pure EV vehicle. In this case, this study raises the problem, PnG energy consumption optimization, to find the most energy-saving PnG operation. What is more, the comparison of energy consumption between PnG and CC is proposed to demonstrate the energy-saving advantage of PnG.

In this problem, the PnG strategy is constrained with a speed fluctuation region $\pm$ 5 km/h, an expected base cruise speed $V_c$ km/h, which means in a PnG cycle, the vehicle firstly accelerates to $V_c$ +5 km/h and then decrease to $V_c$ -5km/h. To find a most efficient PnG operation is to determine best PnG parameters, the appropriate acceleration value $a_{x1}$ and deceleration value $a_{x2}$, to make vehicle save energy as more as possible.

To solve the this problem determining appropriate $a_{x1}$ and $a_{x2}$ and minimizing the energy cost in the PnG operation, this study employs Intelligent Genetic Algorithm and Particle Swarm Optimization (IGPSO), an algorithm combines the advantages of Genetic Algorithm (GA) and Particle Swarm Optimization (PSO), to obtain the optimal solution. GA and PSO are both search algorithms while GA places emphasis on the natural search of optimization and PSO focus on the comparative process search. IGPSO selects the searching process of PSO based on which the advantage of global search of GA is leveraged when good swarms are searching for better swarms in PSO and simultaneously GA could be employed to explore surrounding states. In this way, the opportunities of PSO obtaining optimal swarms could greatly increase [18]. The workflow chart of IGPSO is presented as Fig.5 shows. The implementation steps could be written as:

Step 1. Initialize the initial parameters like the number of optimized parameters, iterations, keeping probability, cross probability, mutation probability, population size, particle dimension.

Step 2. Initialize the initial population based on the region of search.

Step 3. Calculate the fitness value of particle swarms to determine the Individual extremum $p_b$ and global extremum $p_g$ of optimal particle.

Step 4. Increase iterations and determine the number of iterations j.

1) When j is an even number, use GA operator to update the position and speed of particles base on equation (21),

$$child(x_i) = p \times parent_1(x_i) + (1.0 - p) \times parent_2(x_i) \qquad (21)$$

where child ($x_i$) represents the random solution obtained after cross and mutation. P is a random number between 0 to 1 representing the .

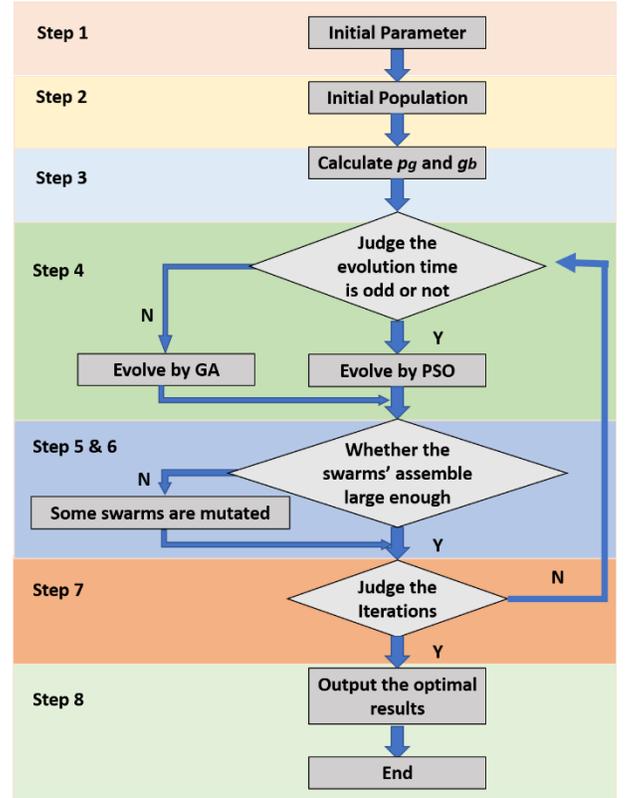

Fig. 5. IGPSO workflow

2) When j is an odd number, use PSO operator and equation (22) and (23) to update the position and speed of particles.

$$v_{id}^{j+1} = w \times v_{id}^j + c_1 r_1 (p_{bid} - z_{id}^j) + c_2 r_2 (p_{gd} - z_{gd}^j) \qquad (22)$$

$$z_{id}^{j+1} = z_{id}^j + v_{id}^{j+1} \qquad (23)$$

In PSO, every individual is regarded as a particle, in a d-dimensions space, every particle is seen as a point. Assuming that the swarm consists of N particles and $z_i = (z_{i1}, z_{i2},…,z_{id})$, $z_i$ is the position vector of i'th particle in d-dimensions. Based on the fitness function, the fitness of $z_i$ could be used to judge the advantages of the position of the particle; $v_i = (v_{i1}, v_{i2},…, v_{id})$ is the speed of i'th particle; $p_{bi} = (p_{bi1}, p_{bi2}, …, p_{bid})$ is the best position of I th particle so far; $P_g = (p_{g1}, p_{g2}, … , p_{gd})$ is the best position of global particles so far. Particles update their position and speed according to equation (22) and (23).

Step 5. According to swarm assemble shown in equation (25), $F_i$ is individual fitness and $F_{avg}$ is the average fitness of the current particle swarm. F is defined in equation (24). if the swarm assemble is over the threshold value then impose mutation process on it shown in equation (26).

$$F = \begin{cases} \max|F_i - F_{avg}|, \max\max|F_i - F_{avg}| > 1 \\ 1, \ else. \end{cases} \quad (24)$$

$$\sigma^2 = \sum_{i=1}^{N} \left(\frac{F_i - F_{avg}}{F}\right)^2 \quad (25)$$

$$z_i = p_b(i) \times (1 + 0.5\mu) \quad (26)$$

Step 6. Again, calculate the fitness value of particle swarms to determine the Individual extremum $p_b$ and global extremum $p_g$ of optimal particle.

Step 7. Determine if iterations meet the requirements. If yes, turn to Step 7. If not, turn to Step 4.

Step 8. Export the global optimal position and optimal solution of the optimal particle.

As a result, the parameter $a_{x1}$ and $a_{x2}$ obtained from the IGPSO computation could make the system generate lowest energy cost after a trip. Besides, the result should be comparted with the energy cost of the CC operation in the same trip to see whether the optimized PnG operation could save more energy than the CC operation. And the energy cost is represented by SOC cost.

### D. The Effect of Regenerative Braking on PnG

The PnG operation practicing on an EV also proposes a problem that how does regenerative braking affect the energy efficiency of PnG operation. Whether or not the regenerative braking could increase the energy cost compared with the optimized PnG operation and the CC operation.

To solve this problem raised above that whether regenerative braking contributes to lowering energy consumption in PnG operation, here imposes the regenerative braking in the PnG operation, and compared its SOC cost with that of the optimal PnG operation and the CC operation.

## VI. SIMULATION RESULTS AND EVALUATION

The simulation study is completed in the combination of Matlab/Simulink and Carsim.

### A. ACC Function implementation

After the simulation, the results of the first scenario are shown as Fig. 6, where the difference of the displacement of two cars have been maintaining a stable state without big fluctuations. From the speed and the inter-car distance sub-figures, the ego car initially accelerates to catch up the leading car, then its speed exceeds the leading vehicle to curtail the inter-car distance, and finally decelerates to the same speed with the leading car with the appropriate inter-car distance. After t = 30 s, the leading car accelerates from 60 to 90 km/h while the ego car accelerates at the preset maximum acceleration 2 m/s$^2$ and finally catches up the leading car.

From the simulation results of the second scenario shown in Fig. 7, the ego car is able to follow the leading car in the first 15 seconds. When another car cut in the lane at t = 15s, the distance between the car cutting in and the ego car is 20 meters which is less than the desired distance according to the spacing strategy, correspondingly, the ego car could decelerate to obtain enough space as the speed sub-figure shows. After that, the ego car could accelerate to catch up the speed of the new leading car and then adjust itself to keep the desired distance to the new leading car, which is the car cutting in, as the speed and inter-car distance figures shows.

From the designed simulation scenarios and their results, this study could verify the implementation of ACC system function as well as the capabilities of following the leading car and executing appropriate actions to keep the desired space according to preset spacing strategy.

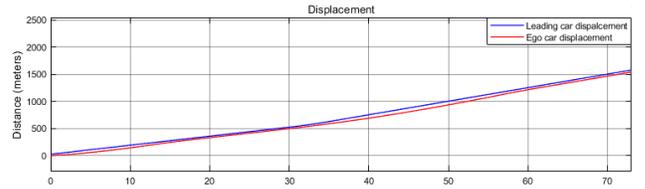

(a) Displacement

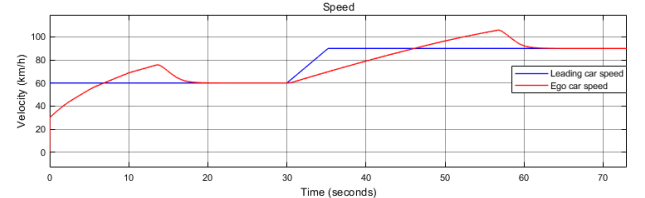

(b) Speed

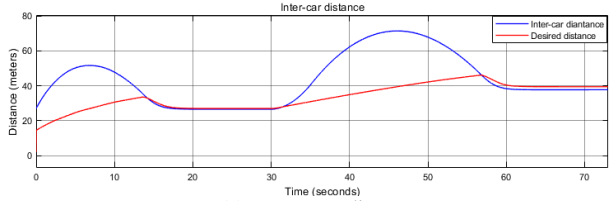

(c) Inter-car distance

Fig. 6. Leading car accelerates

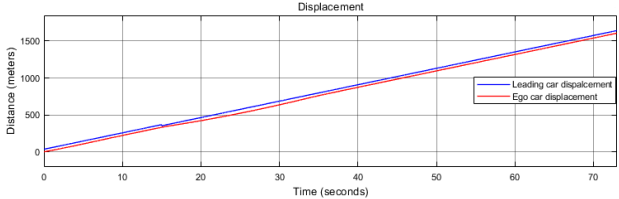

(a) Displacement

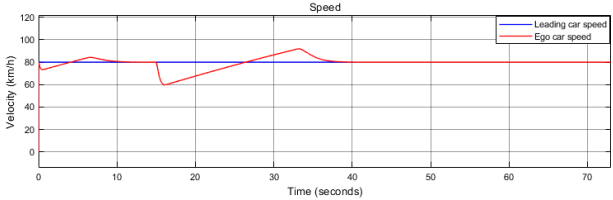

(b) Speed

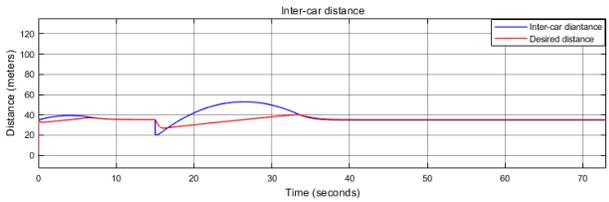

(c) Inter-car distance

Fig. 7. Another car cuts in lane

### B. Regenerative Braking Performance

After simulation, as shown in Fig. 8, where the blue line represents the standard NEDC vehicle speed versus time and the red lines represents that of the ego vehicle executing an NEDC, the vehicle could perform the NEDC with some small perturbations. From Fig. 9, after one NEDC, the SOC of the situation using regenerative braking drops from 0.9 to 0.7559 while the SOC of the simulation using solely hydraulic braking drops from 0.9 to 0.5867. Thus, using regenerative braking could cut down 54% SOC cost compared with using hydraulic braking solely, which illustrates the advantages of regenerative braking on saving-energy in a NEDC test.

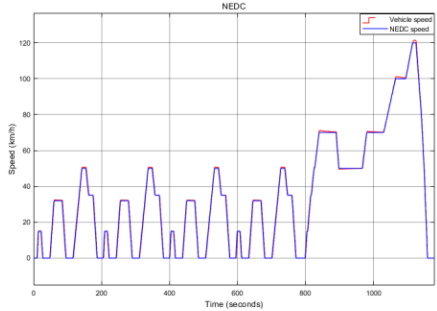

Fig. 8. NEDC speed

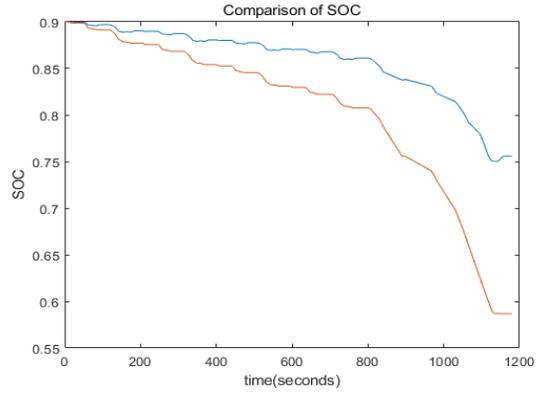

Fig. 9. Comparison of SOC. The blue line represents the SOC of using regenerative braking and the red line represents the SOC of using hydraulic braking solely.

### C. PnG Energy Consumption Optimization

In the simulation study, base cruise speed $V_c$ is set to 30km/h. The vehicle speed region is [25, 35] according to the PnG strategy introduced above. The optimization parameters are acceleration value $a_{x1}$ and deceleration value $a_{x2}$. The $a_{x1}$ is restrained in the region [0, 2] and $a_{x2}$ is restrained in the region [-2, 0], which means the swarm size has just one dimension. The swarm size is set to 20. Max iterations is set to 20. The keep probability is 20%, the cross probability is 40% and the mutation probability is 40%, which is shown as Table III. The fitness function is defined with difference value of SOC cost. The optimization object is minimizing the SOC cost when practicing the PnG operation. The simulation stops at 5 kilometers.

TABLE III.  IGPSO PARAMETERS

| Parameters | Value |
| --- | --- |
| Optimized Objects | [ax1, ax2] |
| Swarm Size | 20 |
| Swarm Dimension | 1 |
| Max Iterations | 20 |
| Cross Probability | 40% |
| Keep Probability | 20% |
| Mutation Probability | 40% |

After practicing the IGPSO algorithm in the simulation study, the optimized values of $a_{x1}$ and $a_{x2}$ are respectively assigned with 0.6122 and -0.0552 m/s$^2$.

As Fig.10 shows, in the optimal PnG operation that obtained from IGPSO computation, the vehicle speed fluctuates between 25km/h to 35km/h. In the PnG state window, 1 represents glide and 0 represents pulse. The SOC cost is optimized to 0.02 which means SOC drops from 0.9 to 0.88 after the 5- kilometers driving. Completing the same 5 kilometers driving at a constant speed of 30 km/h in a CC way, the SOC drops from 0.9 to 0.8721 and the cost of SOC is 0.0279, as shown in Fig.11, the PnG strategy is verified efficacious to cut down the energy consumption by 28.3%.

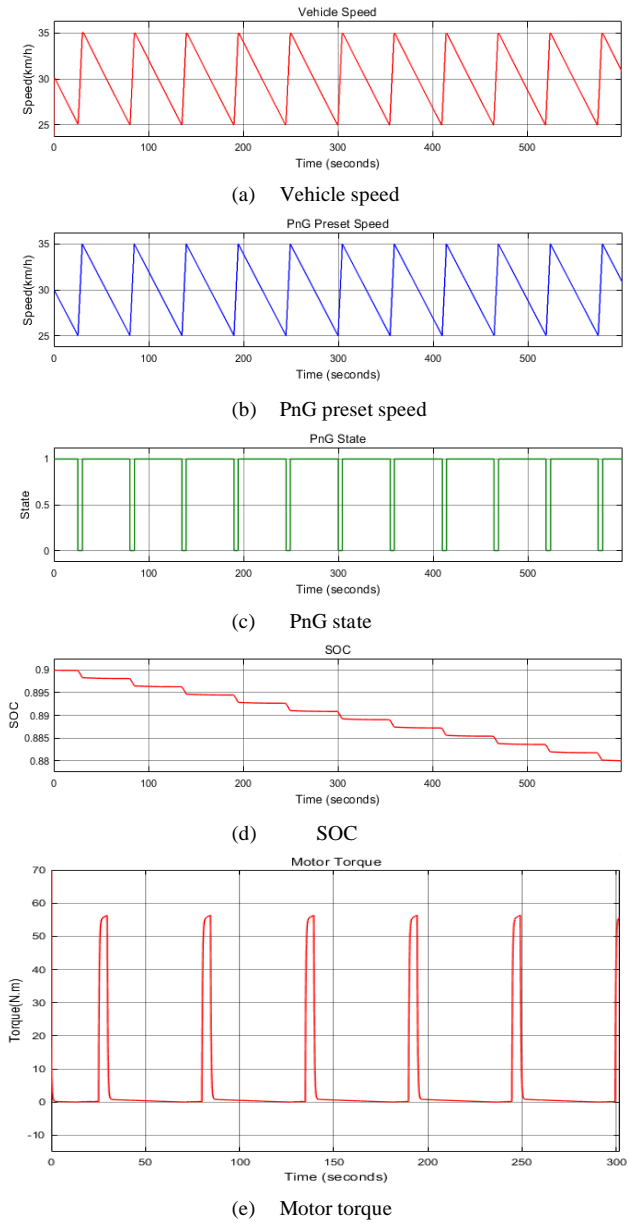

(a) Vehicle speed

(b) PnG preset speed

(c) PnG state

(d) SOC

(e) Motor torque

Fig. 10. Optimal PnG operation

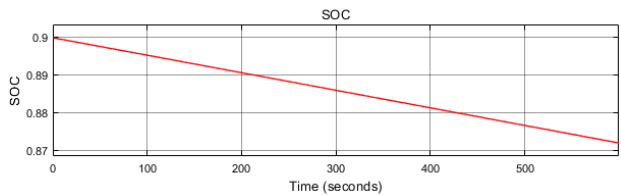

Fig. 11. CC operation

When looking at the SOC curve, there is no SOC recovery which means the regenerative braking is not employed in the optimal solution. The EV decrease its speed by rolling and air resistance with the small positive motor power. This is also shown in motor torque sub-figure in Fig.10, where the motor torque is above 0, which means no braking command executed by motor so that regenerative braking is not triggered and no energy recovered in regenerative braking.

In summary, after the IGPSO computation, the optimized PnG operation could save more energy than the CC operation. Additionally, the PGACCS the model proposed in this paper could be an effective energy-saving model that has the potential to be practiced on EV as a parallel function of CC function when SOC is insufficient.

### D. The Effect of Regenerative Braking on PnG

To solve the problem raised above that whether regenerative braking contributes to lowering energy consumption in PnG operation, here imposes the regenerative braking in the PnG driving that modifies the deceleration value $a_{x2} = -0.5$ to trigger the regenerative braking, then carries out the simulation.

From the results shown in Fig. 12, The cost of SOC is 0.0586 which is bigger than the CC operation and optimal PnG operation. The generative braking is employed which is presented in the SOC and motor torque sub-figures where the negative motor torque means that regenerative braking triggered.

After a series of simulations, once regenerative braking is employed, the energy consumption is bigger than optimal PnG condition where the regenerative braking is not employed and vehicle slides basically without motor power. And in most cases the energy consumption of employing regenerative braking is bigger than the CC operation. This result is mainly caused by the energy loss in the regenerative braking process. Because the transforming of kinetic energy to electrical energy in the regenerative braking process is executed in the motor and battery both of which are not in 100% efficiency.

Due to the energy transforming loss, vehicle itself could keep more energy as the kinetic form than transforming it to the electrical form which also demonstrates the validity of the optimal PnG strategy obtained from IGPSO computation because there is no regenerative braking employed. As a result, regenerative braking contributes less to lowering SOC cost when carrying out the PnG operation.

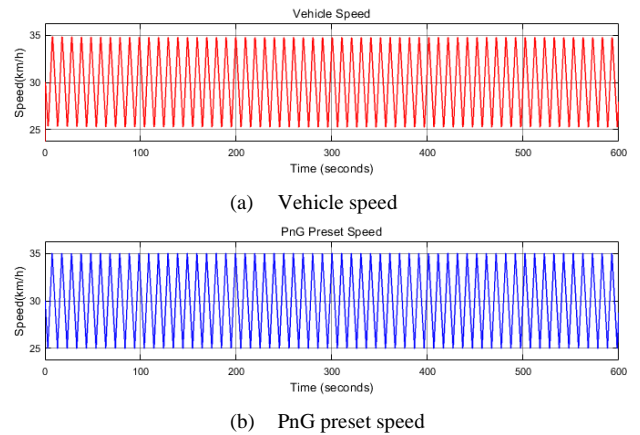

(a) Vehicle speed

(b) PnG preset speed

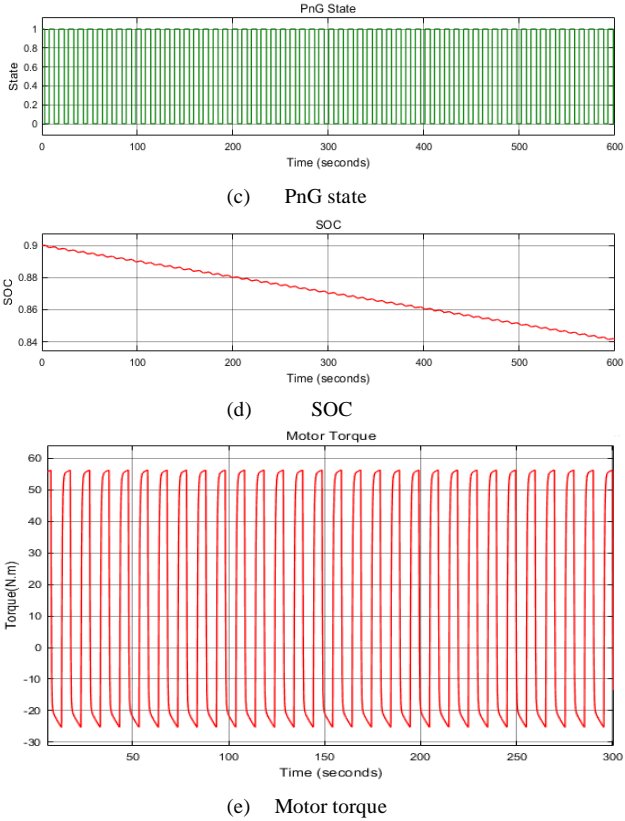

(c) PnG state

(d) SOC

(e) Motor torque

Fig. 12. Regenerative braking PnG operation

## VII. DISCUSSION

From the results of simulation studies presented above, that PnG operation could benefit the lowering the energy consumption. In this section, the prospect and the limitation of this study are discussed. Demonstrated helpful to save more energy compared with cruise control (CC), PnG has its advantages on energy-saving as an embedded function of PGACCS. Meanwhile, the PnG operation is more confined within strict operating conditions. The discussion of these advantages and disadvantages could be in favor of later researches on related fields.

### A. Limitation of This Paper

Although this paper obtains the results that PnG on EV saves more energy than CC, the thing could not be overlooked that the battery model and motor model this paper selected based on which the results are obtained are not universal to every vehicle. What is more, this study has a strong dependency on simulation but not a real vehicle, it confines the demonstration of PnG strategy in the real world. However, the simulation model could be a helpful reference to later researchers. What is more, this study has not discussed driving comfort issue of the PnG operation. In the practical application, when acceleration of pulse phase or deceleration of glide phase are incorrectly set, the Jerk value could be big enough to make driver feel uncomfortable.

### B. Limitation of PnG strategy on EV

PnG operation requires longer safety distance from both leading and following vehicles which means PnG could not easily placed in adaptive cruise control in a ACCS because adaptive cruise control is usually employed in a car-following scenario where the following capacity is the kernel function to track the leading vehicle whereas PnG could trespass the spacing strategy of adaptive cruise control to make system unstable and increase the risk of end-to-end collisions as well as the damper of transportation efficiency. However, when the road condition is good enough for CC, it could also make sense on PnG. As a result, in design, the PnG could not be placed in adaptive cruise control subsystem within an ACCS, it could be placed in cruise control subsystem as a function parallel with cruise control as mentioned in Sec. III. Except for that, PnG also has the limitation in providing driving comfort for drivers.

### C. Prospect of PnG Strategy and PGACCS on EV

As EV becomes more popularized nowadays, there are increasing methods to improve the energy performance of EV. There are some researches about energy-saving on EV. For instance, literature [18] proposed the application of the electric double layer capacitor (EDLC) blocks with the change-over between the series and the parallel modes, to the traction inverter circuit for pure electric vehicles. Literature [19] proposed a novel load management solution for coordinating the charging of multiple plug-in electric vehicles (PEVs) in a smart grid system to diminish power loss in charging process. Literature [20] proposed a fuzzy based safe power management system for EV to increase the effectiveness of energy-saving in terms of speed and acceleration. These researches evaluate the approaches on EV's energy saving from the view of respect of energy reservation unit, charging grid and energy management focusing on mostly hardware level. PnG strategy, however, is an achievable and accessible driving strategy to cut down energy consumption whose advantages lie in the unnecessary improvements on expensive powertrain and components because PnG costs no substantial investments. As this paper obtains, the PnG operation could have higher energy efficiency than the CC operation or a constant speed driving that could cut down 28.3% energy cost in a 5 kilometers trip. Compared to the hardware improvements, manufactures could hopefully employ PnG strategy as an economy mode on EV in the future. In this case this study also shed light on PGACCS that places PnG in an ACCS to parallel it with the system function, cruise control (CC). The benefit of employing PGACCS is that when SOC is insufficient, EV could switch from CC mode to PnG mode to save energy. For the characteristics of EV, insufficient energy density is always a tricky problem. Given that improvements on hardware are restricted, PGACCS would be a low-cost approach to enhance the energy performance of EV. Manufactures could calibrate the PnG base speed and the fluctuation region corresponding to different driving speeds before the vehicle come on the market and provide the PnG mode button that could be triggered as divers' need.

## VIII. CONCLUSIONS

In this study, the electric vehicle model with the PnG-driven adaptive cruise control system and regenerative braking is built up. The four problems are formulated and solved. The first one is the implementation of adaptive cruise control function in the model. The model could practice adaptive cruise control successfully in the two preset operating conditions subject to the preset ACC strategy. The second problem is exploring the regenerative braking effects on saving energy. This study uses a contrast experiment testing the energy consumption on the conditions employing regenerative braking and employing solely hydraulic braking in the NEDC test. The result is that using regenerative braking saves 54% SOC cost compared to using hydraulic braking solely. The third problem is to find the optimal PnG strategy and minimize the SOC cost. Given that the base speed of vehicle is 30km/h and varies from 25 to 35km/h when executing PnG strategy, through the IGPSO computation, this study determines the best value for acceleration value $a_{x1}$ and deceleration value $a_{x2}$ to obtain optimal SOC cost which is 28.3% less than the cruise control operation at 30km/h in a 5 kilometers trip. The fourth problem explores the energy effect of regenerative braking on the PnG operation in which this study finds that the regenerative braking could cost more energy than the optimal PnG solution in which no brake applied, and even more than the constant speed driving due to the energy loss in the transforming process. Based on this study, PnG strategy could be a promising method to enhance the energy performance of EV; PGACCS could be a more energy-saving ACCS for EV.

Besides the abovementioned implementations in this study, PnG strategy could be further discussed in other applications. As mentioned in the discussion section, PnG has a limitation that requires enough safety concerns to practice in some driving scenarios. However, PnG may be helpful in low-speed driving scenarios where the safety concern is less important. One of the low-speed applications is delivery robot, which is a thriving industry in recent years [41]. Therein lies more challenges on energy consumptions increased by the increasing delivery robots or vehicles [40]. In the future studies, PnG strategy could be tested on delivery robots which could be operates in the low-speed scenarios like campuses and sidewalks and have the potential to cut down considerable energy consumptions caused by rapidly growing robotic delivery networks. Additionally, what kind of challenges will PnG strategy bring to power electronics design and stability analysis is also an interesting problem. Due to the length and scope of this study, stability analysis of power electronics is not included. The related power electronics researches about stability control and small signal regulation [42], however, are potentially have effect on PnG implementation process that could be further studied in the future.

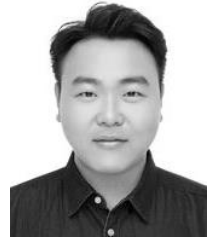

**Zhaofeng Tian** *received the B.S. degree in college of automotive engineering, Jilin University, Changchun, China, in 2019. He is currently working toward the MS degree at Wayne State University, Detroit, MI, USA. His current research interests include electric vehicles, intelligent vehicular robots, and autonomous driving.*

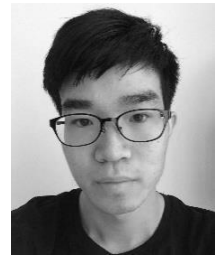

**Liangkai Liu** *received the B.S. degree in telecommunication engineering from Xidian University, Xi'an, China, in 2017. He is currently working toward the Ph.D. degree at Wayne State University, Detroit, MI, USA. His current research interests include edge computing, distributed systems, and autonomous driving.*

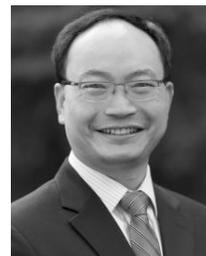

**Weisong Shi** *(F'16) received the B.S. degree in computer engineering from Xidian University, Xi'an, China, in 1995, and the Ph.D. degree in computer engineering from the Chinese Academy of Sciences, Beijing, China, in 2000. He is a Charles H. Gershenson Distinguished Faculty Fellow and a Professor of computer science with Wayne State University, Detroit, MI, USA. His current research interests include edge computing, computer systems, energy-efficiency, and wireless health.*